\documentclass[runningheads]{llncs}
\usepackage[T1]{fontenc}
\usepackage{amsmath,amssymb,amsfonts}
\usepackage{makecell}
\usepackage{algorithmic}
\usepackage{multicol}
\usepackage{multirow}
\usepackage[misc]{ifsym} 
\usepackage{graphicx,verbatim}

\begin{document}
\title{Enjoying Information Dividend: Gaze Track-based Medical Weakly Supervised Segmentation}
\titlerunning{Enjoying Information Dividend}
\author{
Zhisong Wang$^{1}$\thanks{Z. Wang and Y. Ye contributed equally. Corresponding author: Y. Xia.} \and
Yiwen Ye$^{1}$$^{\star}$ \and
Ziyang Chen$^{1}$ \and
Yong Xia$^{1,2}$$^{(\textrm{\Letter})}$
}
\authorrunning{Z. Wang et al.}
\institute{
National Engineering Laboratory for Integrated Aero-Space-Ground-Ocean Big Data Application Technology, School of Computer Science and Engineering, \\
Northwestern Polytechnical University, Xi'an 710072, China \\
\and
Ningbo Institute of Northwestern Polytechnical University, Ningbo 315048, China
\email{yxia@nwpu.edu.cn}
}

\maketitle

\begin{abstract}

Weakly supervised semantic segmentation (WSSS) in medical imaging struggles with effectively using sparse annotations. One promising direction for WSSS leverages gaze annotations, captured via eye trackers that record regions of interest during diagnostic procedures. However, existing gaze-based methods, such as GazeMedSeg, do not fully exploit the rich information embedded in gaze data. In this paper, we propose GradTrack, a framework that utilizes physicians’ gaze track, including fixation points, durations, and temporal order, to enhance WSSS performance. GradTrack comprises two key components: Gaze Track Map Generation and Track Attention, which collaboratively enable progressive feature refinement through multi-level gaze supervision during the decoding process. Experiments on the Kvasir-SEG and NCI-ISBI datasets demonstrate that GradTrack consistently outperforms existing gaze-based methods, achieving Dice score improvements of 3.21\% and 2.61\%, respectively. Moreover, GradTrack significantly narrows the performance gap with fully supervised models such as nnUNet.

\keywords{Eye-tracking \and Gaze Supervision \and Segmentation.}

\end{abstract}
\section{Introduction}
Fully supervised medical image segmentation methods require pixel-wise annotations, which are time-consuming and labor-intensive. As a result, weakly supervised semantic segmentation (WSSS), which leverages sparse annotations, has emerged as a promising alternative to significantly reduce annotation costs. Such sparse annotations include image-level labels \cite{tang2024hunting,zhao2024sfc}, bounding boxes \cite{cheng2023boxteacher,tian2021boxinst}, point annotations \cite{cheng2022pointly}, and scribbles \cite{chen2025addressing,li2024scribformer,luo2022scribble,wang2024few}. Recently, advancements in human-computer interaction technologies \cite{ghosh2023automatic} have enabled physicians to use eye trackers to scan regions of interest (ROIs) during diagnostic procedures. These eye trackers provide gaze-based annotations, marking fixation points along with their respective durations and start timestamps, offering an efficient alternative to traditional annotations without additional effort, thus further reducing annotation costs.
\begin{figure}[t]
    \centering
    \includegraphics[width=1\linewidth]{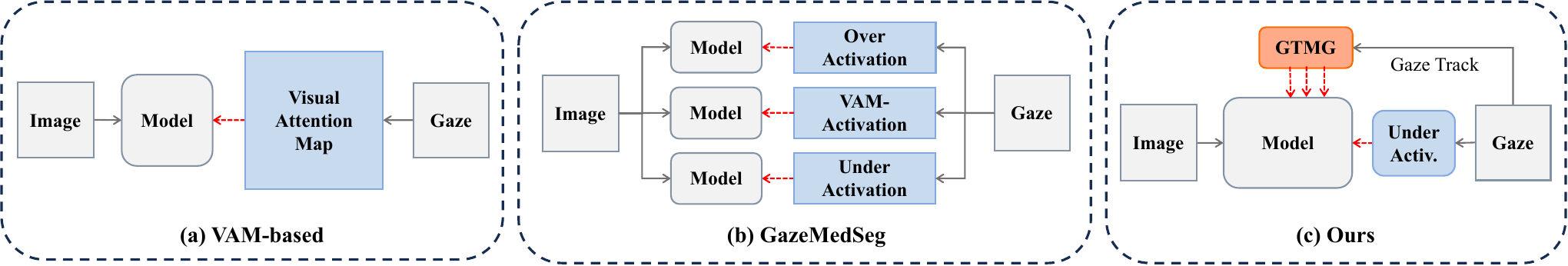}
    \caption{Three frameworks of gaze-based WSSS. (a) VAM-based: Directly using VAM as supervision. (b) GazeMedSeg: Applying multi thresholds to VAM to generate over- and under-activation maps for joint supervision. (c) GradTrack: Using multiple gaze track attention maps to provide stronger information for overly under-activation maps. ``Activ.'': Abbreviation of activation.}
    \label{fig:fig1}
\end{figure}

Existing gaze-based methods focus on utilizing gaze annotations as auxiliary priors to enhance segmentation \cite{jiang2024glanceseg}, classification \cite{wang2024gazegnn}, and detection tasks \cite{kong2024gaze}. 
For segmentation, a common approach is to convert gaze data into a visual attention map (VAM) (see Fig. \ref{fig:fig1}(a)) by associating the duration of fixation points with their level of importance, typically through the convolution of an isotropic Gaussian kernel over the fixation points. Regions around fixation points with longer durations are given higher weights. The heatmap can also be refined using dense conditional random fields (D-CRF) \cite{krahenbuhl2011efficient}. However, the VAM-based approach often introduces noisy and redundant information, leading to performance degradation.
To the best of our knowledge, GazeMedSeg \cite{zhong2024weakly} is the first gaze-based WSSS method in medical image segmentation. It applies multiple thresholds to VAM refined by D-CRF to generate over- or under-activated hard masks and trains multiple expert models to learn from them. While this method helps mitigate noise and redundancy, the limited information derived from gaze annotations still constrains GazeMedSeg’s effectiveness.
We argue that \textbf{both the fixation points and their durations are valuable, but the order of these fixation points, inferred from their start timestamps, must also be considered}. 
During diagnostic imaging, physicians dynamically shift their attention across different ROIs, progressively refining their conclusions. These gaze patterns inherently capture the diagnostic reasoning process, encompassing initial analysis, intermediate cognitive processing, and final decision-making \cite{bisogni2024gaze,ibragimov2024use}.

Motivated by these insights, we propose GradTrack, a \textbf{Grad}ual supervision framework based on gaze \textbf{Track}s for medical image segmentation. In contrast to GazeMedSeg, we use a more conservative thresholding strategy to achieve more reliable results, even at the cost of increased under-segmentation. Additionally, we introduce the gaze track map generation (GTMG) module and track attention (TA) module to guide the model in progressively refining its segmentation predictions. The GradTrack framework is shown in Fig. \ref{fig:fig1}(c). Specifically, the GTMG module generates track maps based on fixation points and their start timestamps. By truncating the gaze sequence from the last fixation point, we create multiple track attention maps, which are then transformed into soft attention maps via distance-based calculations. These soft track maps are treated as supervision signals to be fed into the TA modules located within the decoder to guide the model. This approach enables the model to predict track attention maps and integrate them into the decoding process, providing strong prior information for segmentation. Unlike the physician’s diagnostic workflow, we argue that noise has a greater impact in the early decoding process. Therefore, we reverse the diagnostic procedure, gradually incorporating more uncertain information until a complete track map is constructed. 
To evaluate our GradTrack, we conduct extensive experiments on two public datasets with different modalities: the Kvasir-SEG dataset \cite{jha2020kvasir} for polyp segmentation in endoscopic images, and the NCI-ISBI dataset \cite{bloch2015nci} for prostate segmentation in T2-weighted MRI scans. 

Our contributions are as follows.
(1) We conduct an in-depth analysis of prior works, highlighting the under-utilization of gaze annotations, particularly neglecting the temporal information.
(2) We design the GTMG module to generate multiple track attention maps, which are served as supervision signals to provide guidance. 
We also devise the TA module to subsequently integrate segmentation predictions.
(3) We present a comprehensive evaluation of GradTrack, comparing it with multiple existing WSSS methods, demonstrating the superiority of GradTrack over other methods.

\section{Methods}

\subsection{Overview}
GradTrack adopts a U-Net model~\cite{ronneberger2015u} as the backbone, consisting of an encoder, a decoder, and a segmentation head. For a gaze annotation, GradTrack first generates multiple gaze track attention maps by using the GTMG module. Then, GradTrack uses the TA module to learn to predict these maps using the outputs of the decoder blocks as inputs. The predictions obtained by the TA module are added into the decoding process for providing strong prior information for segmentation. The pipeline of our GradTrack was illustrated in Fig. \ref{fig:overall}. We now delve into the details of each part.

\subsection{Gaze Track Map Generation Module}
\begin{figure}[t]
    \centering
    \includegraphics[width=1\linewidth]{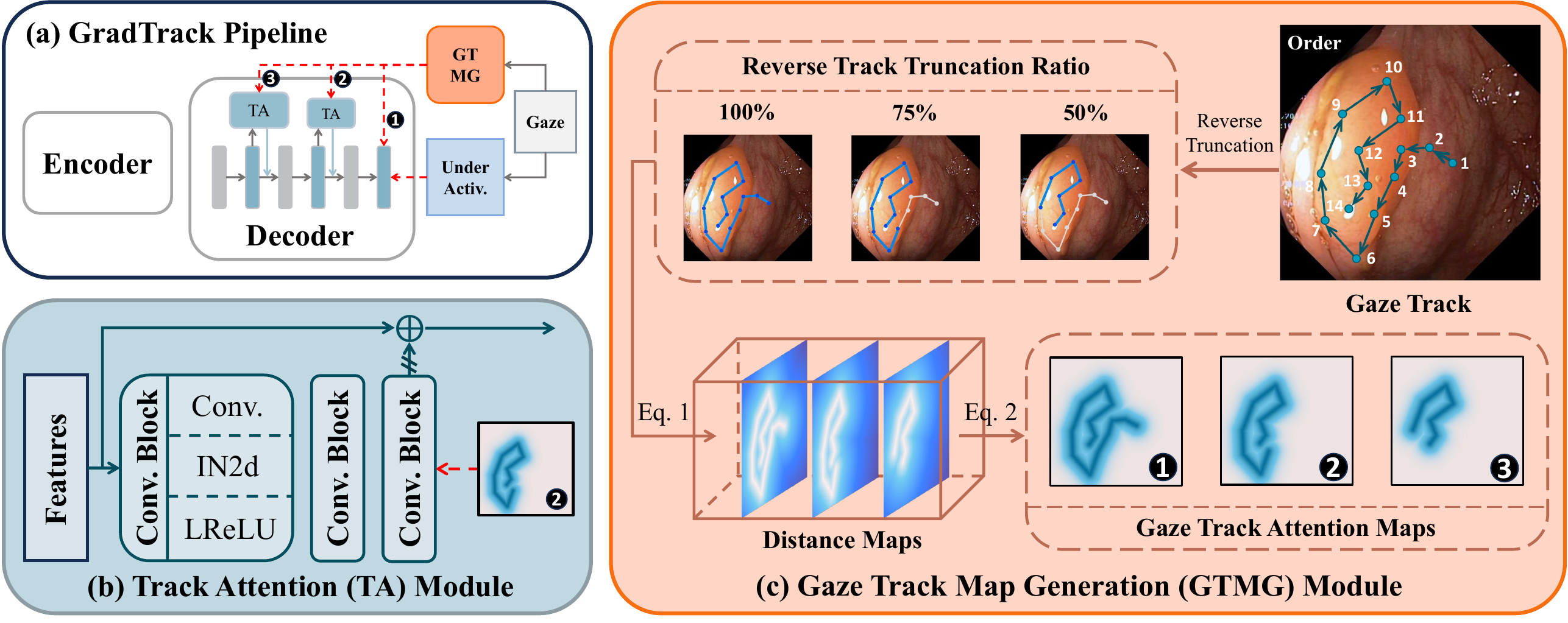}
    \caption{Overview of the proposed GradTrack. (a) The training pipeline of GradTrack, where the red dashed lines denote the supervision information. (b) Track Attention (TA) Module: Composed of three convolutional blocks, it enhances the model’s prior information by truncating the learned GTMG information and fusing it with the main features. (c) Gaze Track Map Generation (GTMG) Module: Generates gaze attention maps by applying different reverse truncation ratios to the gaze track and using a distance-based exponential decay function to enrich the supervision information. ``Activ.'': Abbreviation of activation.}
    \label{fig:overall}
\end{figure}

Generating gaze track attention maps aims to simulate the diagnosis of physicians. As shown in Fig. \ref{fig:overall} (c), we first generate the gaze track $T$ according to fixation points and their start timestamps. Then, we reversely truncate $T$ with 50\%, 75\%, and 100\% ratios, creating $T^{50\%}$, $T^{75\%}$, and $T^{100\%}$, respectively. 
Then, we compute the Euclidean distance from each pixel $p$ to the nearest point $t$ on the gaze track $T^r$ with ratio $r$, resulting in a distance map $D^r$ as follows:
\begin{equation}
D^r(p) = \min_{t \in T^r} \|p - t\|_2.
\end{equation}
Subsequently, we utilize an exponential decay function to convert each $D^r$ into the gaze track attention map $G^r$, which can be is defined as:
\begin{equation}
G^r = e^{-D^r/\beta} \cdot \mathbb{I}(e^{- D^r/\beta} > \tau),
\label{eq:exp}
\end{equation}
where $\beta$ controls the decay rate, $\tau$ represents the field of view threshold, and $\mathbb{I}(\cdot)$ is an indicator function. It ensures that only regions with sufficient confidence are retained in $G^r$.

\subsection{Track Attention Module}
To introduce gaze track priors into the model, we design the TA module. The TA modules are added into the decoder blocks. The output features $F^h$ of the 2nd and 4th decoder blocks are separately processed through a TA module to generate a prediction $P^h$, while the 6th (last) block directly generates $P^h$.
Each $P^h$ consists of two channels, denoting background prediction ($P^h_{bg}$) and foreground prediction ($P^h_{fg}$), where $h$ denoting the block number. The gaze track attention maps generated by the GTMG module are used to supervise the prediction by:

\begin{equation}
\mathcal{L}_{GTMG}^{r,h} = - \Big[(1 + G^r_{re}) \cdot \mathbb{I}{(G^r_{re}>0)} \cdot \log(P^h_{fg}) + \mathbb{I}{(G^r_{re}=0)} \cdot \log(P^h_{bg})\Big],
\end{equation}
where $G^r_{\text{re}}$ denotes the bilinearly interpolated version of $G^r$, resized to match the spatial resolution of $P^h$.

For $h=2 /4$, we extend the channel dimension of $P^h_{fg}$ from $1$ to the one of $F^h$. $F^h$ is further enhanced by $P^h_{fg}$ through element-wise fusion, formulated as: 
\begin{equation}
\hat{F}^h = F^h \oplus |P^h_{fg}|,
\end{equation}
where $\oplus$ means element-wise adding, and $|\cdot|$ is the stop-gradient operation. 

\subsection{Optimization of GradTrack}

For the under-activation hard pseudo ground truths, generated by applying a more conservative threshold (\emph{i.e.}, 0.7) to VAM$_{\text{D-CRF}}$ to preserve a smaller portion of the activation information. 
In contrast, GazeMedSeg employs thresholds of 0.3 and 0.2 for the Kvasir-SEG and NCI-ISBI datasets, respectively.

For the basic overly under-activation hard label supervision, we employ the sum of Dice loss and cross-entropy loss as the supervision mechanism, denoted as $\mathcal{L}_{\text{VAM}}$. The overall loss function integrates all individual losses into a unified optimization objective, formulated as:

\begin{equation}
\mathcal{L} = \mathcal{L}^{100\%,6}_{GTMG} + \lambda_{1} \cdot \mathcal{L}^{50\%,2}_{GTMG} + \lambda_{1} \cdot \mathcal{L}^{75\%,4}_{GTMG} + \lambda_{2} \cdot \mathcal{L}_{VAM},
\end{equation}

where $\lambda_{1}$ and $\lambda_{2}$ are balancing coefficients that control the contribution of each term.
This loss function ensures multi-level optimization, enabling the model to effectively learn from both gaze track information and visual attention mechanism.
 
\section{Experiments and Results}

\subsection{Datasets and Evaluation Metric}

\subsubsection{Datasets.} 

We evaluated the performance of our proposed GradTrack model on two publicly available medical imaging datasets: Kvasir-SEG for polyp segmentation and NCI-ISBI for T2-weighted MRI prostate segmentation. The Kvasir-SEG dataset comprises 900 training images and 100 test images, while the NCI-ISBI dataset includes 789 training images and 117 test images. The training annotations were meticulously generated by annotators with professional training in eye-tracking technology, ensuring high-quality gaze-based supervision for weakly supervised learning.

\subsubsection{Evaluation Metric.} 
The Dice similarity coefficient (Dice) is used to evaluate segmentation performance by measuring the overlap between the predicted segmentation and the ground truth.

\subsection{Implementation Details}

We employ nnU-Net \cite{isensee2021nnu} as the backbone architecture, optimized using Stochastic Gradient Descent (SGD) with a batch size of 8.
The initial learning rate $l_0$ was set to $0.01$ and decayed according to the polynomial rule \cite{isensee2021nnu} $l_t = l_0\times (1-t/T)^{0.9}$, where t is the current epoch and T is the maximum epoch, which was set to 100.
All input images are preprocessed by resizing to $224\times 224$ resolution.
The balancing coefficients $\lambda_1$ and $\lambda_2$ are both set to $0.5$, empirically.
The decay rate $\beta$ is set to 10 for the Kvasir-SEG dataset and 15 for the NCI-ISBI dataset. The field of view threshold $\tau$ is set to 0.25 for both datasets.

\subsection{Results}

\begin{table}[ht]
\centering
\setlength{\tabcolsep}{10pt} 
\caption{Performance comparison of two fully supervised, ten WSSS methods, and our GradTrack on the Kvasir-SEG and NCI-ISBI datasets across three trial seeds. The best results except for fully supervised methods are highlighted in \textbf{bold}.
``Std.'': Abbreviation of standard deviation.}
\begin{tabular}{c|c|c|c|c|c}
\hline\hline
\multirow{2}{*}{\textbf{Method}} & \multirow{2}{*}{\textbf{Supervision}} & \multicolumn{2}{c|}{\textbf{Kvasir-SEG}} & \multicolumn{2}{c}{\textbf{NCI-ISBI}} \\ 
\cline{3-6} 
        &  & Mean & Std. & Mean & Std. \\
\hline

U-Net \cite{ronneberger2015u}          & Full  & 82.12 & 1.11 & 80.58 & 0.48 \\
nnU-Net \cite{isensee2021nnu}          & Full  & 88.41 & 0.47 & 82.43 & 0.18 \\
\hline

PointSup \cite{cheng2022pointly}      & Point & 73.05 & 1.64 & 73.46 & 4.71 \\

AGMM \cite{wu2023sparsely}            & Point & 75.57 & 0.84 & 73.86 & 1.26 \\

AGMM \cite{wu2023sparsely}            & Scribble & 67.23 & 1.02 & 72.70 & 1.03 \\
USTM \cite{liu2022weakly}             & Scribble & 66.31 & 0.93 & 56.89 & 1.23 \\
DMPLS \cite{luo2022scribble}          & Scribble & 69.23 & 0.32 & 57.44 & 0.56 \\
BoxInst \cite{tian2021boxinst}        & Box   & 65.72 & 2.97 & 73.78 & 1.15 \\
BoxTeacher \cite{cheng2023boxteacher} & Box   & 73.33 & 1.30 & 75.60 & 1.15 \\
VAM & Gaze & 72.21 & 0.71 & 73.21 & 0.47\\
VAM$_{\text{D-CRF}}$ \cite{krahenbuhl2011efficient}   & Gaze & 73.12 & 0.60 & 73.86 & 1.91\\
GazeMedSeg \cite{zhong2024weakly}     & Gaze  & 77.80 & 1.02 & 77.64 & 0.57 \\
\hline
GradTrack (Ours)   & Gaze  & \textbf{81.01} & 0.66 & \textbf{80.25} & 0.40 \\
\hline\hline
\end{tabular}
\label{tab:taball}
\end{table}

\subsubsection{Comparison with state-of-the-art WSSS methods.}

We evaluate our model against multiple state-of-the-art WSSS methods, including point-supervised approaches (PointSup \cite{cheng2022pointly}, and AGMM \cite{chen2025addressing}), scribble-supervised methods (USTM \cite{liu2022weakly}, AGMM \cite{chen2025addressing}, and DMPLS \cite{luo2022scribble}), bounding box-supervised methods (Boxinst \cite{tian2021boxinst}, and BoxTeacher \cite{cheng2023boxteacher}), and gaze-supervised methods (VAM, VAM$_{\text{D-CRF}}$ \cite{krahenbuhl2011efficient}, and GazeMedSeg \cite{zhong2024weakly}). 
The scribble annotations in AGMM adopt the style proposed by \cite{valvano2021learning}, while DMPLS and USTM follow the annotation style introduced by \cite{luo2022scribble}. 
For gaze-based methods, VAM and VAM$_{\text{D-CRF}}$ normalize the VAM maps by treating regions with values below 0.5 as background in both the Kvasir-SEG and NCI-ISBI datasets, utilizing these normalized soft labels for supervision. Notably, the results of PointSup, AGMM, Boxinst, BoxTeacher, and GazeMedSeg are inherited from \cite{zhong2024weakly}.
The results shown in Table \ref{tab:taball} reveal that our GradTrack outperforms all gaze-based supervision methods, achieving the highest performance across both datasets. Specifically, it improves the Dice score by 3.21\% on Kvasir-SEG and 2.61\% on NCI-ISBI compared to the best comparing gaze-based approach, \emph{i.e.}, GazeMedSeg. 
Our GradTrack exhibits remarkably competitive performance compared to fully supervised methods, achieving results only 1.11\% and 0.33\% lower than the fully supervised U-Net on the Kvasir-SEG and NCI-ISBI datasets, respectively.
This demonstrates the effectiveness of our GradTrack in narrowing the gap between WSSS methods and fully supervised methods.

\begin{figure}[t]
    \centering
    \includegraphics[width=\linewidth]{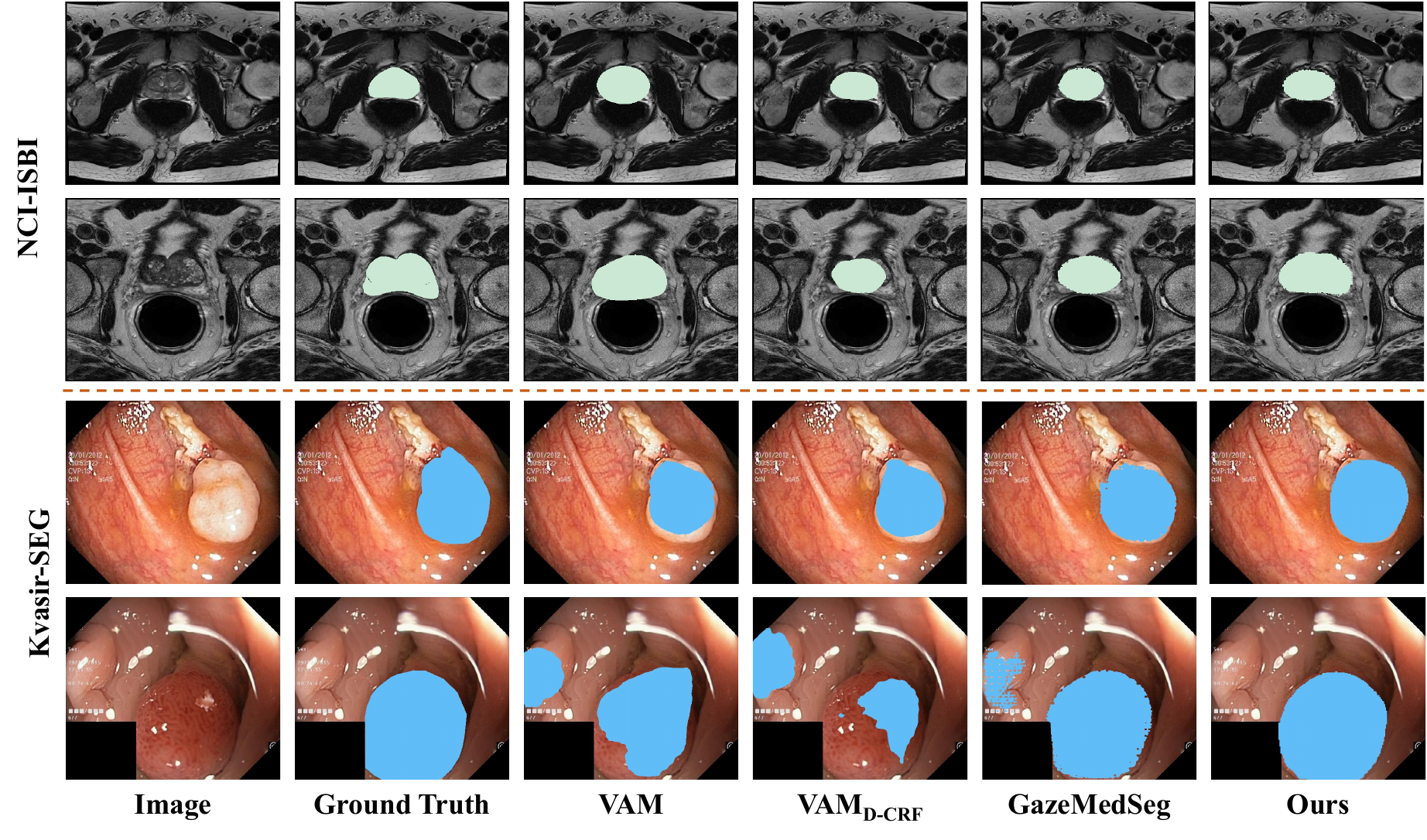}
    \caption{Visualization of segmentation results obtained using gaze-based methods, including VAM, VAM$_{\text{D-CRF}}$, GazeMedSeg, and GradTrack, on the NCI-ISBI and Kvasir-SEG datasets.}
    \label{fig:result}
\end{figure}

\begin{figure}[t]
    \centering
    \includegraphics[width=0.9\linewidth]{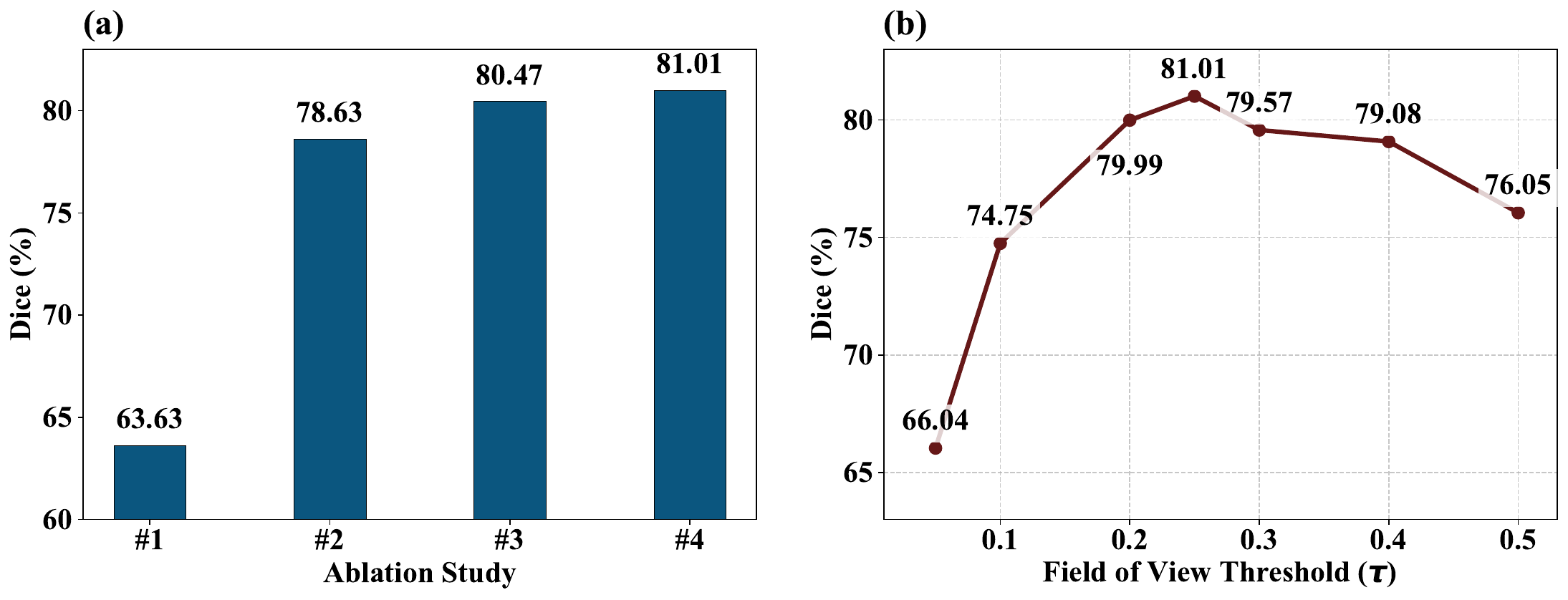}
    \caption{Ablation study and hyper-parameter $\tau$ discussion on the Kvasir-SEG dataset. (a) Analyzing the contribution of each component within our GradTrack. (b) Performance of our GradTrack with various values of $\tau$.}
    \label{fig:ablation}
\end{figure}

\begin{table}[ht]
\centering
\setlength{\tabcolsep}{8pt} 
\caption{Performance ($mean_{std}$) of our GradTrack and four variants on the Kvasir-SEG dataset across three trial seeds. The best result is highlighted in \textbf{bold}. ``Enc.'': Abbreviation of encoder. ``Dec.'': Abbreviation of decoder.}
\begin{tabular}{c|c|c|c|c|c}
\hline\hline
\multirow{2}*{Method} & \multicolumn{2}{c|}{Truncation} & \multicolumn{3}{c}{TA Location} \\
\Xcline{2-6}{0.4pt}
& w/o & Sequential & Enc. + Dec. & Enc. & Dec. (Ours)\\ 
\hline
Dice & $80.28_{0.02}$ & $79.16_{0.19}$ & $80.09_{0.53}$ & $79.96_{0.30}$ & $\textbf{81.01}_{0.66}$ \\ 
\hline\hline
\end{tabular}
\label{tab:strategies}
\end{table}

For qualitative analysis, we select two images from each dataset to compare the segmentation results with other gaze-based supervision methods, alongside the ground truth. The segmentation comparison is shown in Fig. \ref{fig:result}. These results demonstrate that our segmentation results are the closest to the ground truth. Notably, in the second row of the Kvasir-SEG dataset, it can be observed that other methods exhibit overconfidence in segmenting polyps, whereas our model does not produce any erroneous predictions.

\subsubsection{Ablation study} 
We conducted an ablation study to assess the effectiveness of each component. The results are presented in Fig. \ref{fig:ablation}(a), where $\#1$ refers to using only the under-activated VAM as supervision, $\#2$ indicates adding the track attention maps generated by the GTMG module for supervision at the final network layer, $\#3$ extends this by using the TA module, and $\#4$ represents our full design, with two reverse-truncated track weighted maps guiding the TA modules to further enhance the decoder’s ability.
It reveals that (1) using only the under-activated VAM leads to a poor performance; (2) introducing both GTMG and TA modules can improve the segmentation; (3) the best performance is achieved when they are jointly used (\emph{i.e.}, our GradTrack).

In Table \ref{tab:strategies}, we further analyze the impact from two aspects: track attention map truncation and the placement of the TA module, resulting in four variants. The results demonstrate that (1) deeper track information is more beneficial for guiding the model’s learning; (2) more accurate information at deeper network layers is more effective than redundant information; and (3) training the model to learn supervisory information in encoder may have a negative effect on the feature extraction ability of the model.

\subsubsection{Sensitivity to field of view thresholds.}

For the track attention maps, the coverage of each track determines the upper limit of the information the model can absorb. By analyzing the field of view range, we present in Figure \ref{fig:ablation} the model’s performance under different field of view thresholds, including 0.05, 0.1, 0.2, 0.25, 0.3, 0.4, and 0.5. When the threshold is set to 0.25, the model absorbs information most effectively. A threshold as small as 0.05 introduces excessive redundancy, disrupting the model’s perception, while a threshold too large, such as 0.5, exposes the model to insufficient data, hindering its ability to learn useful information.

\section{Conclusion}

We propose GradTrack, a gaze-based weakly supervised segmentation model, consisting of GTMG and TA modules. The GTMG module converts gaze data into the gaze track attention map using reverse truncation and a distance-based exponential decay function, and the TA module provides guidance to the decoder under the multi-level supervision. 
Extensive experiments on the Kvasir-SEG and NCI-ISBI datasets demonstrate superior performance of our GradTrack, which can significantly reduce pixel-level annotation requirements and decrease the gap with fully supervised methods.





\bibliographystyle{splncs04}  
\bibliography{references}    
\end{document}